\documentclass{article}

\usepackage[main, final]{neurips_2026}

\usepackage[utf8]{inputenc} 
\usepackage[T1]{fontenc}    
\usepackage{hyperref}       
\usepackage{url}            
\usepackage{booktabs}       
\usepackage{amsfonts}       
\usepackage{nicefrac}       
\usepackage{microtype}      
\usepackage{xcolor}         

\usepackage{amsthm}
\usepackage{mathtools}
\RequirePackage[capitalize,noabbrev]{cleveref}
\theoremstyle{plain}
\newtheorem{theorem}{Theorem}[section]
\newtheorem{proposition}[theorem]{Proposition}
\newtheorem{lemma}[theorem]{Lemma}

\theoremstyle{definition}

\theoremstyle{remark}

\theoremstyle{plain}
\newtheorem{theorem*}{Theorem}
\usepackage{mynotation}
\usepackage[header]{appendix}
\usepackage{titletoc}

\makeatletter
\AddToHook{cmd/appendix/before}{\def\cref@section@alias{appendix}\def\cref@subsection@alias{appendix}}
\makeatother

\title{Market Incentives for AI Safety Investment}

\author{%
    Nikita Tsoy \\
    INSAIT\\
    Sofia University ``St. Kliment Ohridski''\\
    Sofia, Bulgaria \\
    \texttt{nikita.tsoy@insait.ai} \\
}

\begin{document}

\maketitle

\begin{abstract}
    Despite their practical significance, modern AI systems pose significant
    societal risks, including toxicity, malicious use, and misinformation.
    Their mitigation depends not only on technical feasibility but also on the
    incentives of firms that develop and deploy these systems. We show how
    profit maximization shapes safety investment in an AI supply chain with an
    upstream LLM provider and several competing downstream firms. In our
    game-theoretic model, we derive the market-driven level of safety
    investment and compare it with the levels that maximize industry profit and
    market welfare. We find that downstream competition generally induces
    downstream firms to invest more in safety than the industry-profit-optimal
    level. By contrast, the upstream provider may underinvest relative to these
    benchmarks because it is not fully compensated for the surplus from safety
    investments. Finally, we analyze how different regulatory interventions
    affect the participants' incentives and profits, and the resulting levels
    of model safety.
\end{abstract}

\section{Introduction}

Modern large language models (LLMs) pose significant societal risks, including
toxicity, malicious use, and misinformation \citep{m24h,x25s}. As artificial
intelligence (AI) systems become increasingly prevalent and economically
important, managing these risks has become a central concern in the machine
learning (ML) literature. Although technical methods for AI risk management
constitute a large and important research direction
\citep{g25c,m24h,o22t,b22c}, the economic incentives that determine whether
such methods are developed and deployed have received relatively less
attention.

These incentives, however, shape the behavior of AI ecosystem participants and
hence play a central role in AI governance. Understanding economic incentives
and the resulting market outcomes is crucial for designing effective
regulations that balance model safety and usefulness, especially in common
supply-chain situations that involve upstream providers and downstream adopters
\citep{h25a}. In this case, well-meaning regulations aimed at improving safety
may backfire and lead to lower safety investment levels since participants have
divergent incentives \citep{l25t}.

\paragraph{Contributions}
In this paper, we theoretically analyze market-driven safety investment in an
AI supply chain. We consider a common setting in which one upstream model
provider serves an industry with multiple competing downstream firms and
demonstrate how profit-maximizing incentives might encourage safety investments
by both the provider and the firms. We focus on lower-impact errors that can be
described by neural scaling laws, not catastrophic risks. In our model, the
firms can access the LLM by paying a fee to the provider. The LLM lowers firms'
production costs but also introduces safety-related costs arising from safety
accidents. The risk of these accidents can be reduced through safety
investments, e.g., by collecting additional data. We use standard economic
models of Cournot competition and Nash-in-Nash bargaining to characterize
downstream competition and access fees, and model safety-related costs using
neural scaling laws established in the ML literature.

Our methodological contributions are two-fold. First, we link the benefits of
additional LLM alignment to competition outcomes. This link shows how market
characteristics shape investment incentives and how safety investments affect
market welfare (a notion of welfare centered on economic outcomes). Second, we
suggest an economic- and machine-learning-grounded way to determine fees to the
provider. Specifically, this description shows how the provider benefits from
safety investments even when the provider does not directly suffer reputational
harm from the model's safety accidents.

In our game-theoretic model, we show that competition characteristics strongly
influence equilibrium levels of safety investment. Specifically, we find that
downstream competition generally induces downstream firms to invest more in
safety than the industry-profit-optimal level. By contrast, the upstream
provider may underinvest because it is not fully compensated for the surplus
generated by its safety investments. Additionally, we show that upstream and
downstream regulations that enforce a minimum safety level lead to higher
safety investment. However, strict regulation reduces the participants'
profits, which might slow adoption or cause the model provider to exit the
market.

\section{Related Work}

\paragraph{Safety incentives in AI supply chains}
Our work analyzes an important problem of safety incentives in AI supply
chains. Similarly to us, \citet{l25t} analyze safety investments and
regulations in a market with one model provider and one firm. However, they use
a stylized quadratic-cost model of safety investments and do not directly model
downstream competition and bargaining processes, which leads to a backfiring
effect of weak safety regulation in their model that we do not observe in our
work. Additionally, they do not study how competition affects safety investment
and how regulation affects market welfare. \citet{j25s} analyze entry into the
LLM market, assuming a provider can enter the market only if its model is
sufficiently safe compared to others. While this work also analyzes
safety-related incentives, it focuses on barriers to entry rather than on the
market incentives for safety investments for incumbent firms.

\paragraph{Economic analysis of AI regulation}
Our work contributes to a growing literature on AI regulation, which includes
several important topics. \citet{l24f} study the investment incentives in
AI supply chains, similarly to our work. However, they do not examine
regulations, link AI investments to market characteristics, or incorporate
consumer utility in their analysis. \citet{c25r,f21g} study audits of AI
systems.  \citet{w25n,q25m,x25e} analyze the effects of AI models' openness.
\citet{h25a} discuss regulatory challenges in AI supply chains. \citet{t25s}
analyze how generative AI affects high-quality data generation. Finally, a
growing body of economic work \citep[e.g.,][]{a24t,g25h} examines the optimal
rate of AI adoption, given that AI risks are not fully known.

\paragraph{Economics of ML}
Our work also belongs to a broader literature on the economics of ML, which
studies data as a new production asset employed by modern ML systems
\citep{s25g}. This literature includes data-sharing incentives \citep{c23d},
data valuation \citep{l25a}, and strategic incentives for data manipulation
\citep{h16s,k22a}. Most closely, our work is related to \citet{t23s}, who study
data-sharing incentives between competing firms. They use the same model of
downstream market competition and a similar model of ML impact on production
costs, but do not consider investment incentives or AI supply chains.

\paragraph{Economics of information security}
Our work is also related to the literature on information security
\citep{m24s}, as incentives in AI safety can be structurally similar to those
in information security. Here, our work is closely related to the work of
\citet{g02t}. Specifically, our parametric form of safety-related costs
(\cref{eq:costs}) is mathematically identical to one of the parametric forms
suggested by \citet{g02t} to describe the risk of security breaches, even
though the motivation for this parametric form differs materially. This
identity makes our \cref{thm:priv-invest} similar to some of \citet{g02t} 's
results. However, the results of \citet{g02t} do not establish the relationship
between optimal investments and market characteristics.  Additionally, our
approach to safety-related costs resembles ex post liability in the information
security literature \citep{m24s}, while the regulations we study resemble ex
ante safety regulation.

\paragraph{AI risk measurement}
While we use an abstract notion of safety, deploying AI systems requires
developing metrics and evaluation protocols to measure AI risk. See
\citet{a24a,g24a,g26s} for an overview of risk metrics and risk frameworks.

\paragraph{Industrial organization}
Our approach to market competition closely follows the standard approaches in
industrial organization \citep{t88t}. Specifically, we use a widely employed
model of Cournot competition with quasilinear quadratic utility consumers
\citep{d79a}. Additionally, to model bargaining between the model provider and
the firms, we use the notion of Nash-in-Nash bargaining, widely used in applied
and theoretical economic work \citep{c19n}.

\section{Framework}

In this section, we introduce a game-theoretic framework that describes safety
investments in AI supply chains. We focus on a common situation in which a
provider offers its foundation model as a service to several competing
downstream firms.\footnote{We briefly discuss an extension of our framework to
the case of several foundation model providers in \cref{sec:many-providers}.}
This model streamlines the firms' operations but also introduces safety-related
costs. These costs arise from safety accidents (e.g., unqualified advice,
toxicity) that lead to reputational losses or regulatory penalties. To reduce
these costs and increase profits, both the provider and the firms invest in
model safety, e.g., by collecting additional data.

As an example, consider several competing travel agencies that can optimize
their operations by introducing an AI chatbot for customer relations. However,
the LLM might exhibit unsafe behavior, e.g., giving unqualified legal advice
when the user asks about the terms of booking cancellation, leading to
reputational losses. Thus, the firms are willing to either train a safer model
themselves using their private data or pay the provider for additional safety
training on synthetic or public data.

\paragraph{Objectives}
The provider and the firms maximize their profits, which have the following
structure:
\begin{equation}
    \label{eq:provider-costs}
    \varPi_p = \sum_i \text{AccessFee}_i - \text{InferenceCosts} -
    \text{ProviderSafetyInvestments}.
\end{equation}
\begin{equation}
    \varPi_i = \text{Sales}_i - \text{ProductionCosts}_i -
    \text{SafetyRelatedCosts}_i - \text{AccessFee}_i -
    \text{FirmSafetyInvestments}_i.
\end{equation}
Here, $\text{AccessFee}_i$ is the fee for using the model paid by firm $i$ to
the provider, $\text{InferenceCosts}$ is the costs of model inference,
$\text{ProviderSafetyInvestments}$ is the provider's safety investments,
$\text{Sales}_i$ is the sales of firm $i$, $\text{ProductionCosts}_i$ is the
production costs of firm $i$, $\text{SafetyRelatedCosts}_i$ is the
safety-related costs incurred by firm $i$, and $\text{FirmSafetyInvestments}_i$
is firm $i$'s safety investments.

Note that the considered structure assumes that only downstream firms incur
safety-related costs. This assumption reflects the difficulty of assigning
direct responsibility for model behavior to the foundation model provider,
given that the provider does not fully control how downstream firms approach
additional safety training and model deployment. Specifically, we assume that
the model provider does not incur direct reputational losses and is not liable
for safety-related accidents. \cref{sec:upstream-liability} discusses an
extension of our framework in which the provider becomes liable for safety
accidents and shows that this case is similar to the case above.

\paragraph{Actions}
Now, we describe the participants' actions. First, both the provider and firms
can invest in safety. These investments might take the form of additional data
collection, data generation, additional alignment training, etc. We denote the
provider's and firm $i$'s investment levels by $n$ and $k_i$, respectively.
Second, each firm and the provider bilaterally bargain over the model access
fee, $t_i$. Third, each firm makes downstream production decisions, $q_i$.

These actions occur in the following order.
\begin{enumerate}
    \item The provider enters the market and chooses its safety investment
        level, $n$.
    \item Firms simultaneously bargain with the provider over access fees,
        $t_i$.
    \item Each firm chooses its level of additional safety investments, $k_i$.
    \item The firms compete in the market by making their production decisions,
        $q_i$.
\end{enumerate}
At each stage, the participants maximize their profits, given the outcomes of
all previous stages.

\subsection{Game Stages}

We now discuss each stage of the game in more detail. Consistent with the
backward induction logic, we discuss each stage in time-reversed order.

\paragraph{Competition}
Classic competition models describe how each firm's sales depend on the production decisions of \emph{all} firms and how each firm's production costs depend on its own production decisions. Additionally, since higher production volume leads to more LLM usage, our framework introduces two new elements that could depend on production decisions: safety-related costs and access fees.

Taken together, the firms optimize the following market profit functions:
\[
    \max_{q_i} \varPi_i^m \defeq \text{Sales}_i(\vec{q}) -
    \text{ProductionCosts}_i(q_i) - \text{SafetyRelatedCosts}_i(q_i, k_i, n) -
    \text{AccessFee}_i(q_i, t_i).
\]
Since sales depend on the production decisions of all firms, we assume that the
resulting decisions, $q_i^*(\vec{k}, \vec{t}, n)$, constitute a Nash
equilibrium.  We denote the resulting market profits by
$\varPi^{m,*}_i(\vec{k}, \vec{t}, n)$.

\paragraph{Firms' investments}
To choose safety investment levels, the firms optimize their profits
\begin{equation}
    \label{eq:invest}
    \max_{k_i} \varPi^{m,*}_i(\vec{k}, \vec{t}, n) -
    \text{FirmSafetyInvestments}_i(k_i).
\end{equation}
We denote the resulting firms' profits by $\varPi_i^*(\vec{t}, n)$, the
resulting safety investment level by $k_i^*(\vec{t}, n)$, and the resulting
provider's profits by
\[
    \varPi_p^*(\vec{t}, n) = \sum_i \text{AccessFee}_i(q_i^*, t_i) -
    \text{InferenceCosts}(\vec{q}^*) -
    \text{ProviderSafetyInvestments}(n).
\]

\paragraph{Bargaining}
To determine the access fees, we use the Nash-in-Nash bargaining solution. This
concept assumes the provider and each firm negotiate bilaterally over several
rounds. At each even round, the provider suggests a fee level. The negotiating
firm can accept or refuse this level, thereby limiting the fee size. At each
odd round, the firm suggests a fee level, and the provider can either accept or
refuse it \citep[see][for game-theoretic foundation]{c19n}. This process
results in the Nash bargaining solution \citep{b86t} for each negotiating pair,
taking the outcomes of other negotiations as given. The resulting fees should
constitute a Nash equilibrium.

Formally, denote $\vec{t}_{-i} \defeq (t_1, \dots, t_{i-1}, t_{i+1}, \dots,
t_m)$, a vector of fees without $i$'s entry. According to the Nash-in-Nash
concept, each pair maximizes the following objective:
\begin{equation}
    \label{eq:bargain}
    \max_{t_i} (\varPi_i^*(\vec{t}, n) - \varPi_{i,-i}^*(\vec{t}_{-i}, n))^{1 -
    \phi} (\varPi_p^*(\vec{t}, n) - \varPi_{p,-i}^*(\vec{t}_{-i}, n))^\phi,
\end{equation}
where $\phi$ is the provider's bargaining power and $\varPi_{i,-i}^*$ and
$\varPi_{p,-i}^*$ are firm $i$'s and the provider's profits if all firms but
$i$ have access to the LLM. We denote the resulting fees by $t_i^*(n)$.

\paragraph{Provider's investments}
Finally, the provider optimizes profits to choose its safety investment level
\[
    \max_{n} \varPi_p^*(\vec{t}^*(n), n).
\]
We denote the resulting investment level by $n^*$.

\section{Game-Theoretic Model}

To obtain analytical results, we focus on a specific instantiation of the
framework. In the following sections, we describe the analytic form of each
component in greater detail.

\subsection{Production-Related Elements}

This section describes production-related elements: competition model,
inference costs, and access fees. We focus on the classical Cournot model of
competition of \citet{d79a}, similarly to \citet{t23s}. This model, widely used
in economic studies \citep{c19q}, describes a market with $m$ firms, where firm
$i$ produces good $i$. We assume that $m \ge 2$. Each firm maximizes its
profits by tailoring its production volume to consumer demand. To describe this
model, we need to specify demand (i.e., consumer behavior) and supply (i.e.,
firm behavior) factors.

\paragraph{Demand}
There are $M$ representative consumers (i.e., $M$ is the market size), each
representing the average behavior of all consumers. Their utility is a
quasi-linear quadratic utility function \citep{c19q}, which they maximize given
market prices and a budget constraint
\begin{equation}
    \max_{\vec{q}, g} u(\vec{q}, g) = \sum_{i=1}^m q_i - \Bigl(\frac{\sum_i
    q_i^2 + \gamma \sum_{i \neq j} q_i q_j}{2}\Bigr) + g \quad \text{s.t.}
    \quad \vec{q}^\tran \vec{p} + g \le G,
\end{equation}
where $q_i$ is the consumed amount of good $i$, $g$ is the remaining money,
$\vec{p}$ is the vector of prices for each good, $G$ is the budget of the
consumer, and $\gamma \in (0, 1)$ is the substitutability between the firms'
goods. Higher $\gamma$ corresponds to more similar goods and, hence, ``more
intense'' competition.

The first-order conditions for this problem yield $m$ demand equations that
link prices and quantities.

\paragraph{Supply}
Firm $i$ produces $M q_i$ units of good $i$, resulting in market profits
\[
    \varPi^m_i = M q_i p_i - \text{ProductionCosts}_i(q_i) -
    \text{SafetyRelatedCosts}_i(q_i, k_i, n) - \text{AccessFee}_i(q_i, t_i).
\]

We assume that production costs are linear in $q_i$ and in the number of tokens
required for production. Additionally, we assume that the number of
safety-related accidents grows linearly with the number of tokens
\citep[see][for a similar idea]{i25w}. These assumptions result in the
following costs:
\begin{equation}
    \text{ProductionCosts}_i(q_i) + \text{SafetyRelatedCosts}_i(q_i, k_i, n) =
    M q_i c_i(k_i, n),
\end{equation}
where $c_i$ includes both production and safety-related costs.

\paragraph{Inference costs and access fees}
Assuming that inference costs grow linearly in tokens,\footnote{See \cref{sec:nonlinear-inference} for the discussion of nonlinear inference costs.} we get
\begin{equation}
    \text{InferenceCosts} = M \Bigl(\sum_i q_i\Bigr) w.
\end{equation}
Under the same assumption, we consider two access fee types: token- and subscription-based,
\begin{equation}
    \text{AccessFee}_i = M q_i t_i \text{ (token-based)}, \quad
    \text{AccessFee}_i = M q_i w + t_i \text{ (subscription-based)}.
\end{equation}

Denote $c^\text{eff}_i \defeq c_i + t_i$ and $c^\text{eff}_i \defeq c_i + w$ in token- and subscription-based cases, respectively. We get
\[
    \varPi^m_i = M q_i (p_i - c^\text{eff}_i) \text{ (token-based)}, \
    \varPi^m_i = M q_i (p_i - c^\text{eff}_i) - t_i \text{
        (subscription-based)}.
\]
Token-based fees effectively increase the production costs, reducing production and token use by the firms. By contrast, subscription-based fees do not affect the production decisions. This logic parallels the standard comparison between marginal and lump-sum taxes \citep{s15e}.

\paragraph{Equilibrium profits}
In Cournot competition, the firms decide how much to produce to maximize their
profits. Given these quantities, prices are determined by the demand equations
so that markets clear (i.e., quantities produced equal quantities demanded).
The following standard lemma \citep{a17o}, specialized to our setting,
characterizes the resulting equilibrium profits.
\begin{lemma}
    \label{lem:profits}
    Assume that $\forall i \, (2 - \gamma) (1 - c_i^\text{eff}) \ge \gamma
    \sum_{j \neq i} (c_i^\text{eff} - c_j^\text{eff})$ and $G$ is sufficiently
    big (e.g., $G \ge m$). Then, firm $i$'s equilibrium profits follow
    \[
        \varPi_i^{m,*}(\vec{c}^\text{eff}, t_i) = M (q^*_i)^2 \text{
            (token-based)}, \quad \varPi_i^{m,*}(\vec{c}^\text{eff}, t_i)
        = M (q^*_i)^2 - t_i \text{ (subscription-based)},
    \]
    where $q^*_i = \frac{2 - \gamma - (2 + \gamma (m - 2)) c_i^\text{eff} +
    \gamma \sum_{j \neq i} c_j^\text{eff}}{(2 - \gamma) (2 + \gamma (m - 1))}$.
\end{lemma}
The first assumption ensures that the consumer demands each good. The second
one ensures that the consumer has enough money to spend. As we can see, firm
$i$'s profits decrease in its production costs and increase in the production
costs of competitors; $\gamma$ exacerbates this dependency.

\subsection{Safety-Related Elements}
\label{subsec:safety}

Now, we consider safety-related costs and safety investments. In our work, we
focus on safety-related properties that can be improved through additional data
collection or generation, along with additional training. These properties
include all properties that can be improved through reinforcement learning from
human feedback (RLHF) \citep{o22t,b22c}, adversarial training
\citep{m24h,d25m}, or supervised fine-tuning \citep{b24s}, such as avoiding
toxicity, refusing to give unqualified advice, and refusing to assist with
misconduct.

\paragraph{Safety-related costs}
We assume that firms have symmetric costs\footnote{We make costs symmetric to
study the model provider's investments analytically. Without this assumption,
the provider's investments behave similarly in qualitative terms but become
less interpretable; see \cref{sec:asym-cost}.} and that, before the
introduction of LLM in the production process, they equal
\begin{equation}
    c_i = a_0.
\end{equation}
The LLM access decreases production costs but introduces safety-related costs
that grow linearly with the number of safety accidents. We assume that the
probability of accidents depends polynomially on the number of data points used
for safety training, consistent with the learning-theory logic \citep{t04o} and
neural scaling laws \citep{h22a}. So, the final costs follow
\begin{equation}
    \label{eq:costs}
    c_i = a + b / (n^\text{eff}_i(k_i, n))^\beta,
\end{equation}
where $n^\text{eff}_i$ corresponds to the effective number of data points used
for safety training of firm $i$'s model, $\beta > 0$ corresponds to learning
easiness, $a$ is the sum of production and safety-related costs for a perfectly
trained Pareto-optimal model, and $b$ corresponds to the safety penalty due to
generalization error.\footnote{Our analysis will be qualitatively similar if we
use a different form of the cost function; see \cref{sec:diff-cost}.} From an
economic perspective, $b$ describes the sensitivity of cost to safety
investments.

In our model, greater safety investments will not decrease the LLM's utility.
The incentive for cost minimization already determines the desired balance
between safety and utility (via $a$) and accounts for the alignment tax
\citep{o22t}. Additional data collection only reduces the generalization error
in finding the cost-minimizing point and, hence, can only be beneficial
\citep{s25gen}.\footnote{Using the results of \citet{s25gen}, we can also
extend the framework to analyze several safety properties (see
\cref{sec:many-cost}). This extension will behave similarly to the one
presented in the main text.} To achieve this balance in practice, one could use
a replay buffer of utility-enhancing data or techniques for alignment tax
mitigation \citep{n26m,s26s}.

\paragraph{Investment costs}
Both the provider and the downstream firms invest in safety. The provider
performs upstream training, which benefits all firms. Each firm performs
downstream training, which benefits only that firm. We do not assume a specific
training pipeline but require that the total pipeline cost, including data
collection and training, scales linearly with the amount of training data.

The procedure might be the following. First, the provider aligns the model
using generated data. Then, the downstream firms further align the model using
their private data, possibly with different training procedures. Our
requirement allows each participant to use replay buffers of utility-enhancing
data whose sizes scale proportionally with the amount of safety data. The data
may be collected through human labeling, RLAIF \citep{b22c}, red-teaming
\citep{p22r}, or other procedures.

We interpret safety investment levels, $n$ and $k_i$, as the number of points
collected for safety training. Since the firms may collect more relevant data
for their tasks, we consider their data more effective. Specifically, following
the results of \citet[][Remark 2]{h19o}, we assume
\begin{equation}
    n^\text{eff}_i = k_i + (n + 1)^\rho,
\end{equation}
where $\rho \in (0, 1)$ corresponds to the transferability of the provider's
data to the firms' tasks. We use $n + 1$ instead of $n$ to represent the
model's baseline safety capabilities before additional training. To highlight
economic meaning, we often refer to $n^\text{eff}_i$ as the cumulative level of
safety investments. To simplify the notation, we use the following change of
variables in the rest of the paper:
\[
    n + 1 \mapsto n.
\]

Our assumptions about the training pipeline yield the following investment
costs:
\begin{equation}
    \text{ProviderSafetyInvestments} = \lambda_p (n - 1), \quad
    \text{FirmSafetyInvestments}_i = \lambda k_i.
\end{equation}
We allow $\lambda_p \neq \lambda$ since the provider's and the firms' training
pipelines may differ.

\subsection{Investment Problems and Outcomes}

We can now derive the precise safety investment problems the participants face.

\paragraph{Absence of disruption}
\cref{eq:invest}, given \cref{lem:profits} and \cref{eq:costs}, results in analytically intractable expressions. To get an analytical answer, we assume
\[
    (2 - \gamma) (1 - a - w) \gg (2 + \gamma (m - 2)) (b + \delta),
\]
where $\delta \defeq a_0 - a - w$ (LLM cost advantage). This condition has two
implications. First, a small $\delta$ implies that access to the LLM does not
allow one firm to drive all competitors out of the market, i.e., the LLM is not
disruptive.  Second, a small $b$ implies that the penalty for lacking
additional safety data is not severe. Note that this condition implies the
first condition in \cref{lem:profits}.\footnote{We briefly discuss how to relax
this assumption and model firm exit in \cref{sec:disruption}.}

We also assume that the introduction of LLM benefits the production process and
is profitable for all participants. In particular, this assumption implies that
there exists a feasible vector of fees $\vec{t}$ that makes the left- and
right-hand sides in \cref{eq:bargain} non-negative.

\paragraph{Approximation of Equations (\ref{eq:invest}) and (\ref{eq:bargain})}

The first-order approximation of market profits leads to the following
approximation of \cref{eq:invest} (see \cref{subsec:priv-approx}):
\begin{equation}
    \label{eq:priv-approx}
    \max_{k_i} -\frac{A b}{(k_i + n^\rho)^\beta} - \lambda k_i, \
    A \defeq \frac{2 (2 + \gamma (m - 2)) M q^*}{(2 - \gamma) (2 + \gamma (m -
    1))},\
    q^* \defeq \frac{1 - a - w}{2 + \gamma (m - 1)}.
\end{equation}
Here, $A$ is the sensitivity of a firm's profits to its production cost, and
$q^*$ denotes the production level with the optimal model. Notice that this
equation implies that each firm's private investment does not depend on the
private investments of other firms.

Now, we consider \cref{eq:bargain}. We get the following firms' profits:
\[
    \varPi^*_i = \varPi_i^{m, *}(\vec{c}^\text{eff}, t_i) - \lambda k_i, \quad
    \varPi^*_{i,-i} = \varPi_i^{m, *}(\vec{c}^\text{eff}_{-i}(a_0), 0),
\]
where $\vec{c}^\text{eff}_{-i}(x) \defeq (c_1^\text{eff}, \dots,
c_{i-1}^\text{eff}, x, c_{i+1}^\text{eff}, \dots, c_m^\text{eff})$, a vector of
effective costs with $i$th entry changed. We also get the following provider's
profits:
\[
    \begin{aligned}
        \varPi^*_p =& \sum_j \text{AccessFee}_j(q^*_i(\vec{c}^\text{eff}), t_j)
        - M w \sum_j q^*_j(\vec{c}^\text{eff}) - \lambda_p (n - 1),\\
        \varPi^*_{p,-i} =& \sum_{j \neq i}
        \text{AccessFee}_j(q^*_j(\vec{c}^\text{eff}_{-i}(a_0)), t_j) - M w
        \sum_{j \neq i} q^*_j(\vec{c}^\text{eff}_{-i}(a_0)) - \lambda_p (n -
        1).
    \end{aligned}
\]

Thus, \cref{eq:bargain} becomes
\[
    \max_{t_i} (\varPi_i^m(\vec{c}^\text{eff}, t_i) - \lambda k_i -
    \varPi_i^m(\vec{c}^\text{eff}_{-i}(a_0), 0))^{1 - \phi}
    (\varPi^*_p - \varPi^*_{p,-i})^\phi.
\]
And its first-order approximation is (see \cref{subsec:approx})
\begin{equation}
    \label{eq:approx}
    \argmax_{t_i} \Bigl(A \delta - \frac{A b}{(k_i + n^\rho)^\beta} - \lambda
    k_i - B (t_i - w)\Bigr)^{1 - \phi} (t_i - w)^\phi,
\end{equation}
where $B = A$ for token-based fees and $B = M q^*$ for subscription-based fees.
For subscription-based fees, $t_i$ is redefined so that $\text{AccessFee}_i = M
q^* t_i$.

\paragraph{Outcomes}
In our analysis, we track several key metrics: cumulative level of safety
investments ($n^\text{eff}_i$), industry profit, and market welfare.  Industry
profit is defined as the sum of participants' profits,
\begin{equation}
    \varPi^\text{ind} \defeq \varPi_p + \sum_i \varPi_i.
\end{equation}
Market welfare is defined as the sum of consumer utility and industry profits,
\begin{equation}
    W^m \defeq M u^* + \varPi^\text{ind}, \text{ where } u^* \defeq
    u(\vec{q}^*, G - \sum_i p_i^* q_i^*).
\end{equation}
Here, $u^*$ is the utility of a representative consumer after competition.
\textbf{Note} that market welfare does not directly account for the harms to
the consumer from safety accidents.\footnote{We consider a notion of welfare
that accounts for harm to consumers in \cref{sec:welfare}.}

\section{Safety Investments}

This section analyzes market-driven investment levels and compares them with
industry- and market-welfare-optimal ones.

\paragraph{Firms' investments}
Market-driven firms' investments follow the following proposition.

\begin{proposition}
    \label{thm:priv-invest}
    The firms' level of investments follows the following formula:
    \[
        k^* = \max(0, k_m - n^\rho), \ k_m \defeq (\beta b A / \lambda)^{1 / (1
        + \beta)}.
    \]
\end{proposition}
$k^*$ is non-increasing in $n$, while $k^* + n^\rho$ is non-decreasing in $n$.
In other words, although firms decrease investment in response to an increase
in the provider's investment, this decrease does not reduce the cumulative
investment levels.

\paragraph{Provider's investments}
\cref{eq:approx} and \cref{thm:priv-invest} give (see \cref{subsec:provider})
\begin{equation}
    \label{eq:provider}
    \varPi_p^* = \frac{m \phi M q^*}{B} \Bigl(A \delta - \lambda k_m - \frac{A
    b}{\max(n^\rho, k_m)^\beta} + \lambda \min(n^\rho, k_m)\Bigr) - \lambda_p
    (n - 1).
\end{equation}
The LLM adoption is profitable for all participants if and only if $\varPi_p^*
\ge 0$. Also note that, as predicted, the subscription-based fees yield larger
profits since they correspond to the smaller $B$.

\begin{theorem}
    \label{thm:invest}
    Denote $n_s \defeq \bigl(m \frac{\phi M q^*}{B} \frac{\rho
    \lambda}{\lambda_p}\bigr)^{1 / (1 - \rho)}$ and $n_b \defeq \bigl(m
    \frac{\phi M q^*}{B} \frac{\rho \beta b}{\lambda_p} A\bigr)^{1 / (1 + \rho
    \beta)}$. The provider's and
    cumulative levels of investments follow the following formulas:
    \[
        n^* = \max(1, \min(n_b, n_s)), \ k^* + (n^*)^\rho = \max(n_b^\rho,
        k_m, 1).
    \]
    $n_s$ is increasing in $\phi$, $\lambda$, and $m$, is decreasing in
    $\lambda_p$, and does not depend on $b$, $a + w$, and $M$. $n_b$ is
    increasing in $\phi$, $b$, $M$, and $m$, is decreasing in $\lambda_p$ and
    $a + w$, and does not depend on $\lambda$. $k_m$ is increasing in $b$ and
    $M$, is decreasing in $\lambda$, $a + w$, and $m$, and does not depend on
    $\phi$ and $\lambda_p$. The dependence of $k_m$ on $\gamma$ is
    non-monotonic, decreasing for small $\gamma$ and increasing for big
    $\gamma$. For subscription-based fees, $n_s$ does not depend on $\gamma$
    and $n_b$ depends non-monotonically (similarly to $k_m$). For token-based
    fees, $n_s$ and $n_b$ are decreasing in $\gamma$.
\end{theorem}

\cref{tab:statics} summarizes the comparative statics. The investments are
generally non-decreasing in the provider's bargaining power ($\phi$),
sensitivity of costs to investments ($b$), and market size ($M$). The
investments are generally non-increasing in data collection costs ($\lambda$
and $\lambda_p$) and non-optimizable costs ($a + w$). At the same time, the
dependence on competition-related factors, $m$ and $\gamma$, is non-monotonic.

\begin{table}[h]
    \caption{Comparative statics}
    \label{tab:statics}
    \begin{tabular}{l l l l}
\toprule
        & Firms' level, $k^*$ & Provider's level, $n^*$ & Cumulative level,
        $k^* + (n^*)^\rho$\\
\midrule
        Non-decreasing & $\lambda_p$, $b$, $M$ & $\phi$, $\lambda$, $b$,
        $M$, $m$ & $\phi$, $b$, $M$, $m$ for big $m$
        \\
        Non-increasing & $\phi$, $\lambda$, $a + w$, $m$ & $\lambda_p$, $a + w$
        & $\lambda$, $\lambda_p$, $a + w$, $m$ for small $m$\\
\bottomrule
    \end{tabular}
\end{table}

We also observe several interesting facts. First, the provider can invest so
much that firms do not invest at all if $m$ is large enough. This happens
because the provider's investments benefit all firms' safety (i.e., they
exhibit a positive externality), so the return on the provider's investment
scales with $m$. Second, stronger downstream competition might lower cumulative
investments, $k^* + (n^*)^\rho$, since $k_m$ and $n_b$ decrease in $\gamma$ for
small $\gamma$, and $k_m$ decreases in $m$. While downstream competition
induces competition in safety investments among the firms (as we show below) it
also lowers the firms' profits, inhibiting their ability to invest in general.
Third, the provider invests more under subscription-based fees. Thus,
subscription fees not only increase the provider's profits but also increase
safety investments.

\paragraph{Industry profit and market welfare}
Now, we derive the optimal investment levels for industry profits and market
welfare when participants coordinate their investments and fees.

\begin{proposition}
    \label{prop:priv-opt}
    The industry, $k_i^\text{ind}$, and welfare, $k_i^\text{mw}$, optimal
    firms' investments follow
    \[
        k^x_i(n) = \max(0, k^x_m - n^\rho), \
        k^x_m \defeq (\beta b A^x / \lambda)^{1 / (1 + \beta)}, \
    \]
    where $x \in \{\text{ind}, \text{mw}\}$ and
    \[
        A^\text{ind} \defeq \frac{2 M q^*}{2 + \gamma (m - 1)},\
        A^\text{mw} \defeq \frac{(3 + \gamma (m - 1)) M q^*}{2 + \gamma (m - 1)}.
    \]
\end{proposition}
The optimal firms' investments are generally lower than the equilibrium ones.
We can see this by analyzing the behavior of the parameter $k_m$, which drives
firms' investments. We get $A^\text{ind} < A$ and, hence, $k^\text{ind}_m \le
k_m$. If $\gamma^2 (m - 1) + \gamma > 2$, we get $A^\text{mw} < A$ and, hence,
$k^\text{mw}_m \le k_m$. This result shows that competition in investments
incentivizes the firms to invest more in safety. We also note that
$k^\text{ind}_m < k^\text{mw}_m$ since market welfare also accounts for the
consumers' utility.

\begin{proposition}
    \label{prop:prov-opt}
    If the firms' investments follow $k_i(n) = \max(0, k_0 - n^\rho)$ and $k_0
    \ge k_m^x$, the provider's optimal investments are
    \[
        n^x = \max(1, n_e^x), \
        n_e^x \defeq
        \begin{cases}
            n_s^\text{opt}, & (n_s^\text{opt})^\rho < k_0,\\
            n_b^x, & (n_b^x)^\rho > k_0,\\
            k_0^{1/\rho}, & \text{otherwise},\\
        \end{cases}
    \]
    where $n_s^\text{opt} \defeq \bigl(m \frac{\rho
    \lambda}{\lambda_p}\bigr)^{1 / (1 - \rho)}$ and $n_b^x \defeq \bigl(m
    \frac{\rho \beta b}{\lambda_p} A^x\bigr)^{1 / (1 + \rho \beta)}$.
\end{proposition}
We observe two opposite effects. On the one hand, the absence of bargaining
increases the investments (note that the multiplier associated with the
presence of bargaining satisfies $\phi M q^* / B < 1$). On the other hand, the
absence of competition in investments decreases them ($A^x < A$ for large
$\gamma$).

The comparison between equilibrium and industry-optimal investment levels
reveals a collective-action problem. When the industry-optimal provider's
investments are higher than the equilibrium ones, the firms are collectively
willing to compensate the provider for the additional investments. However, due
to the nature of fee bargaining, this compensation does not naturally occur.

\section{Effects of Regulation}

In this section, we discuss how regulatory actions can impact our outcome
metrics. We assume perfect enforcement of regulations. Specifically, we assume
that one can implement a properly designed audit procedure (with sufficiently
large fines) that corresponds to fixed administrative costs for the regulator
and does not change incentives for safety investments.

\paragraph{Downstream regulation}
First, we consider regulations that enforce the safety of the resulting
downstream models. Assuming enforcement focuses on the probability of safety
accidents, it will result in the constraint $n^\text{eff}_i \ge k_0$.

\begin{proposition}
    \label{prop:downstream}
    Consider the downstream regulation $n^\text{eff}_i \ge k_0$. If $k_0 \le
    k_m$, the investment levels will not change. If $k_0 \ge k_m$, we get
    \[
        k^* = \max(0, k_0 - n^\rho), \
        n^* = \max(1, n_e), \text{ where }
        n_e = 
        \begin{cases}
            n_s, & n_s^\rho < k_0,\\
            n_b, & n_b^\rho > k_0,\\
            k_0^{1/\rho}, & \text{otherwise}.\\
        \end{cases}
    \]
    And, we have $n^* \ge n^{*, \text{old}}$, where $n^{*, \text{old}}$ is the
    investment level in no regulation regime.
\end{proposition}
The regulation does not reduce the participants' investments. However, the
firms incur additional data-collection costs and may refuse to use the LLM
under overly strict regulation (see \cref{eq:provider} with $k_m \mapsto k_0$).
The effect on industry profit and market welfare is ambiguous. On the one hand,
regulation increases firms' investment, which reduces these metrics under
severe competition, as shown in \cref{prop:priv-opt}. On the other hand,
regulation also increases the provider's investment. These opposing effects
imply that regulation increases the metrics when $k_0 \le \max(k_m^x,
\min(n_s^\rho, (n_b^x)^\rho))$. In this case, the metrics increase either by
\cref{prop:priv-opt} or because firms make zero investments and benefit from
higher provider investment. Otherwise, the regulation might have a negative
effect.

\paragraph{Upstream auditing}
Another approach is to audit the provider's model, essentially requiring $n \ge
n_0$.

\begin{proposition}
    \label{prop:upstream}
    Under the upstream regulation $n \ge n_0$ ($n_0 \ge 1$), the provider
    investment level will be $n^* = \max(n_0, \min(n_s, n_b))$ and the firms'
    response will not change, $k^* = \max(0, k_m - n^\rho)$.
\end{proposition}
The regulation does not reduce the cumulative level of investments due to
\cref{thm:priv-invest}. However, again, overly strict regulation might force
the provider to exit the market. For industry profit and market welfare, when
$k_m^x \le k_m$, the regulation is beneficial if $n_0 \le n^x$ by
\cref{prop:prov-opt}. Specifically, when the firms make non-zero investments,
the moderate regulation is always beneficial since $n_s < n_s^\text{opt}$. When
the firms make zero investments, the regulation is beneficial if the market is
not overly competitive or the provider's bargaining power is low.

\paragraph{Indirect regulations}
The regulator might also increase safety investments through less direct
actions. \cref{thm:invest} suggests several channels: safety penalty, $b$, and
investment costs, $\lambda$ and $\lambda_p$. By increasing fines for safety
accidents, the regulator indirectly increases $b$ (and possibly $a$, but in
this discussion we assume that the increase in $a$ is small). This increase
raises cumulative investments (through $k_m$ or $n_b$). However, as previously,
this intervention reduces the participants' profits. By subsidizing safety
investments, the regulator decreases $\lambda$ or $\lambda_p$. If the firms
make non-zero investments, the subsidy to the firms leads to higher cumulative
investments (through $k_m$). Otherwise, the subsidy to the provider leads to
higher cumulative investments (through $n_b$). However, both subsidies require
additional funds from the regulator. Also note that the subsidy to the provider
might help address the collective action problem identified in
\cref{prop:prov-opt}, allowing the regulator to finance part of the subsidy
through higher taxes on participants.

\section{Conclusion}

We analyzed how profit maximization might incentivize safety investments by the
participants of the AI supply chain. In our framework, downstream firms are
interested in safer models because they incur safety-related costs. At the same
time, the provider is interested in safety because it can charge higher access
fees to the firms. We characterized the market-driven level of safety
investments and compared it with the industry-profit- and
market-welfare-optimal ones. This comparison shows that downstream competition
generally leads to higher levels of safety investment.

We outlined several ways the regulator could increase safety investment. Each outlined approach has upsides and downsides. Auditing and fines do not require additional funds and lead to higher safety investments but reduce profits from LLM adoption and, hence, might slow the adoption. Subsidies lead to higher investment levels and profits but require additional funds from the regulator.

We hope that our work contributes to AI governance by highlighting the economic
challenges associated with AI safety investment for both policymakers and
industry.

\paragraph{Limitations}
While we demonstrate how profit maximization incentivizes safety investment and
obtain analytical expressions for its levels, our analysis has several
limitations. First, we assume that the firms are symmetric and, hence, have
similar production processes. This assumption might not hold in industries with
substantial horizontal differentiation among firms. Second, our analysis does
not account for catastrophic risks posed by the models. Instead, we focus on
lower-impact errors that can be described by neural scaling laws. Third, our
results rely on the absence-of-disruption assumption, which might not hold in
more dynamic and innovative industries. Fourth, we assume perfect enforcement
in our analysis of regulations and assume that one can design a proper audit
procedure by imposing sufficiently large fines. Finally, we assume specific
functional forms for the elements of our framework. These forms capture the
main conceptual properties of the elements but might differ from real-world
relationships.

\paragraph{Future directions}
We foresee several possible extensions of our model. First, a more general
model might analyze asymmetric firms, which is important in horizontally
differentiated industries. As a related point, it might be interesting to
consider cascading effects in which firms drive competitors out of the market
by gaining access to the LLM. Finally, it might be interesting to consider a
more complex provider market, which might include open-weight models or several
competing providers, as we discuss in \cref{sec:many-providers}.

\section*{Acknowledgments}

This research was partially funded by the Ministry of Education and Science
of Bulgaria (support for INSAIT, part of the Bulgarian National Roadmap for
Research Infrastructure). The authors thank Nikola Konstantinov for helpful
comments on the paper's positioning and writing.

\bibliography{ta-bib}

\clearpage

\appendix

\appendixpage

\startcontents
\printcontents{}{1}[2]{}

\section{Proofs}

\subsection{Proof of Lemma \ref{lem:profits}}

\paragraph{Demand}
Assuming that the restrictions $q_i \ge 0$ and $p_i \ge 0$ will hold in the
optimum, the consumer problem results in the following Lagrange function:
\[
    L = \sum_{i=1}^m q_i - \Bigl(\frac{\sum_i q_i^2 + \gamma \sum_{i \neq j}
    q_i q_j}{2}\Bigr) + g + \nu (G - \sum_{i=1}^m p_i q_i - g).
\]
Karush-Kuhn-Tucker conditions give
\[
    \begin{cases}
        0 = \pdv{L}{q_i} = 1 - q_i - \gamma \sum_{j \neq i} q_j - \nu p_i,\\
        0 = \pdv{L}{g} = 1 - \nu,\\
        0 = \nu (G - \sum_{i=1}^m p_i q_i - g),\\
    \end{cases}
\]
where $\nu$ is the Lagrange multiplier. This system results in the following
equations:
\begin{equation}
    \label{eq:price}
    p_i = 1 - q_i - \gamma \sum_{j \neq i} q_j.
\end{equation}
Notice that $p_i \le 1$ and $q_i \le 1$, hence, $\sum_i p_i q_i \le m$. Thus,
for $G \ge m$, the budget constraint holds. Since the function $u(\vec{q}, g)$
is strongly concave in $\vec{q}$ and linear in $g$, this solution is a global
maximizer.

\paragraph{Supply}
Given that price will be determined by \cref{eq:price}, the firms get the
following production surplus:
\[
    M (p_i q_i - c_i^\text{eff} q_i) = M (1 - c_i^\text{eff} - q_i - \gamma
    \sum_{j \neq i} q_j) q_i.
\]
Thus, the best responses of the firms follow
\[
    0 = \pdv{\varPi^m_i}{q_i} = 1 - c_i^\text{eff} - \gamma \sum_{j \neq i} q_j
    - 2 q_i.
\]
In Nash equilibrium, all actions should follow the best response, resulting in
the system
\[
    q_i^* = \frac{1 - c_i^\text{eff} - \gamma \sum_{j \neq i} q_j^*}{2}.
\]
Therefore,
\[
    \sum_i q_i^* = \frac{m - \sum_i c_i^\text{eff} - \gamma (m - 1) \sum_i
    q_i^*}{2} \implies \sum_i q_i^* = \frac{m - \sum_i c_i^\text{eff}}{2 +
    \gamma (m - 1)}.
\]
By substituting the sum back,
\[
    q_i^* = \frac{2 - \gamma - (2 + \gamma (m - 2)) c_i^\text{eff} + \gamma
    \sum_{j \neq i} c_j^\text{eff}}{(2 - \gamma) (2 + \gamma (m - 1))}.
\]

Notice that $\varPi^m_i = M (x q_i - q_i^2)$ for token-based fees, where $x$ is
some constant that does not depend on $q_i$. Thus, $q^*_i = x/2$ and
$\varPi^{m, *}_i = M x^2 / 4$, resulting in $\varPi^{m, *}_i = M (q_i^*)^2$.
The expression for the subscription-based fees can be derived similarly.

\paragraph{Non-negativity}
Finally, we need to verify that, in equilibrium, we get $q_i^* \ge 0$ and
$p_i^* \ge 0$. For $q_i^*$, we get
\[
    q_i^* \ge 0 \iff 2 - \gamma - (2 + \gamma (m - 2)) c_i^\text{eff} + \gamma
    \sum_{j \neq i} c_j^\text{eff} \ge 0 \iff (2 - \gamma) (1 - c_i^\text{eff})
    \ge \gamma \sum_{j \neq i} (c_i^\text{eff} - c_j^\text{eff}).
\]
For $p_i^*$, notice that the best response quantity for the firm is positive
only if $p_i \ge 0$. However, we showed that $q_i^* \ge 0$ under the
restriction on costs; hence, $p_i^* \ge 0$ under the same restriction.

\subsection{Derivation of Equation (\ref{eq:priv-approx})}
\label{subsec:priv-approx}

The main idea of this approximation is to use a first-order Taylor expansion for all formulas around the point $\vec{c} = (a + w, a + w, \dots, a + w)$.

First, notice that the existence of a fee vector implies $\delta \ge 0$; otherwise, the introduction of the technology is not profitable for either a firm or the provider.

Now, we consider the case of token-based fees. In this case, any feasible fee vector will satisfy $\forall i \ t_i \in [w, \delta + w]$. Under this restriction, the effective marginal production costs are close to the optimal cost $a + w$. We get
\[
    0 \le c^\text{eff}_i - a - w \le \max\Bigl(\delta,
    \frac{b}{(n^\text{eff}_i)^\beta} + t_i - w\Bigr) \le \max(\delta, b + t_i -
    w) \le b + \delta.
\]
Thus, we can use the first-order approximation of profits
\begin{equation}
    \label{eq:profit-approx}
    \varPi^{m, *}_i(\vec{c}^\text{eff}) \approx M (q^*)^2 - \frac{2 (2 + \gamma
    (m - 2)) M q^* (c^\text{eff}_i - a - w)}{(2 - \gamma) (2 + \gamma (m - 1))}
    + \frac{2 \gamma M q^* \bigl(\sum_{j \neq i} (c^\text{eff}_j - a -
    w)\big)}{(2 - \gamma) (2 + \gamma (m - 1))},
\end{equation}
where $q^* \defeq \frac{1 - a - w}{2 + \gamma (m - 1)}$.

Similarly, in the case of subscription-based fees, we get
\[
    0 \le c^\text{eff}_i - a - w \le \max\Bigl(\delta,
    \frac{b}{(n^\text{eff}_i)^\beta}\Bigr) \le \max(\delta, b) \le b + \delta.
\]
And we get the following profits:
\[
    \varPi^{m, *}_i(\vec{c}^\text{eff}) \approx M (q^*)^2 - \frac{2 (2 + \gamma
    (m - 2)) M q^* (c^\text{eff}_i - a - w)}{(2 - \gamma) (2 + \gamma (m - 1))}
    + \frac{2 \gamma M q^* \bigl(\sum_{j \neq i} (c^\text{eff}_j - a -
    w)\big)}{(2 - \gamma) (2 + \gamma (m - 1))} - t_i.
\]

In both cases, \cref{eq:invest} results in the following problem:
\[
    \max_{k_i} - \frac{A b}{(k_i + n^\rho)^\beta} - \lambda k_i,
\]
where $A \defeq \frac{2 (2 + \gamma (m - 2)) M q^*}{(2 - \gamma) (2 + \gamma (m
- 1))}$.

\subsection{Derivation of Equation (\ref{eq:approx})}
\label{subsec:approx}

For token-based fees, the market surplus of LLM introduction follows
\[
    \varPi^{m, *}_i(\vec{c}^\text{eff}) - \varPi^{m,
    *}_i(\vec{c}^\text{eff}_{-i}(a_0)) \approx A (a_0 - a - w) - A
    (c^\text{eff}_i - a - w) = A \Bigl(\delta -
    \frac{b}{(n^\text{eff}_i)^\beta} + w - t_i\Bigr).
\]
As for the profits of the provider, note
\[
    \text{AccessFee}_i(q_i, t_i) - M q_i w  = M q_i (t_i - w) = M q^* (t_i - w)
    + M (q_i - q^*) (t_i - w) \approx M q^* (t_i - w).
\]
Thus,
\[
    \varPi^*_p - \varPi^*_{p,-i} \approx M q^* (t_i - w).
\]
Putting two approximations together, we get the following approximation of Equations (\ref{eq:bargain}) and (\ref{eq:invest}):
\[
    \argmax_{t_i} \Bigl(A \delta - \frac{A b}{(k_i + n^\rho)^\beta} - \lambda
    k_i - A (t_i - w)\Bigr)^{1 - \phi} (t_i - w)^\phi, \
    \argmax_{k_i} - \frac{A b}{(k_i + n^\rho)^\beta} - \lambda k_i.
\]

For subscription-based fees, we have
\[
    \text{AccessFee}_i(q_i, t_i) - M q_i w = t_i.
\]
Thus, we get the following approximation of Equations (\ref{eq:bargain}) and
(\ref{eq:invest}):
\[
    \argmax_{t_i} \Bigl(A \delta - \frac{A b}{(k_i + n^\rho)^\beta} - \lambda
    k_i - t_i\Bigr)^{1 - \phi} t_i^\phi, \
    \argmax_{k_i} - \frac{A b}{(k_i + n^\rho)^\beta} - \lambda k_i.
\]
We can get an equivalent problem by replacing $t_i \mapsto M q^* (t_i - w)$
\[
    \argmax_{t_i} \Bigl(A \delta - \frac{A b}{(k_i + n^\rho)^\beta} - \lambda
    k_i - M q^* (t_i - w)\Bigr)^{1 - \phi} (t_i - w)^\phi, \
    \argmax_{k_i} - \frac{A b}{(k_i + n^\rho)^\beta} - \lambda k_i.
\]
Since $M q_i w \approx M q^* w$, this change of variables gives
\[
    \text{AccessFee}_i \approx M q^* w + t_i \mapsto M q^* t_i.
\]

\subsection{Proof of Proposition \ref{thm:priv-invest}}

We get the following Lagrange function:
\[
    L = -\frac{A b}{(k_i + n^\rho)^\beta} - \lambda k_i + \nu k_i,
\]
where $\nu$ is the Lagrange multiplier. The first-order conditions give
\[
    \begin{cases}
        0 = \pdv{L}{k_i} = \frac{\beta A b}{(k_i + n^\rho)^{\beta + 1}} -
        \lambda + \nu,\\
        0 = \nu k_i,\\
        k_i \ge 0, \: \nu \ge 0.
    \end{cases}
\]
In the case $k_i = 0$, we get
\[
    \frac{\beta A b}{n^{\rho (\beta + 1)}} - \lambda = -\nu \le 0.
\]
Otherwise,
\[
    \frac{\beta A b}{(k_i + n^\rho)^{\beta + 1}} - \lambda = 0 \implies
    \frac{\beta A b}{n^{\rho (\beta + 1)}} - \lambda \ge 0.
\]
Thus, we get
\[
    k_i^* = k^* = \max(0, k_m - n^\rho), \ k_m \defeq \Bigl(\frac{\beta A
    b}{\lambda}\Bigr)^{1 / (1 + \beta)}.
\]

For the cumulative level, we get
\[
    k_i^* + n^\rho = \max(k_m, n^\rho).
\]
Both $k_m$ and $n$ are non-decreasing in $n$; hence, the cumulative level is
non-decreasing in $n$. For $n^\rho \le k_m$, the cumulative level equals $k_m$,
which does not depend on $n$.

\subsection{Derivation of Equation (\ref{eq:provider})}
\label{subsec:provider}

\cref{thm:priv-invest} gives the following expression for the surplus of the
firm:
\[
    \varPi^{m, *}_i - \lambda k_i^* - \varPi^{m, *}_{i, -i} = A \delta -
    \frac{A b}{\max(k_m, n^\rho)^\beta} - \lambda \max(k_m - n^\rho, 0) - B
    (t_i - w) \eqdef S - B (t_i - w).
\]
Then the Nash bargaining concept reduces to
\[
    \max_{t_i} (S - B (t_i - w))^{1 - \phi} (t_i - w)^\phi.
\]
The condition for the existence of feasible fees and the LLM's profitability
for the firms is
\[
    S \ge 0 \iff 
    A \delta - \lambda k_m - \frac{A b}{\max(n^\rho, k_m)^\beta} + \lambda
    \min(n^\rho, k_m) \ge 0.
\]
If the feasibility condition holds, the first-order condition is
\[
    -\frac{(1 - \phi) B}{S - B (t_i - w)} + \frac{\phi}{t_i - w} = 0,
\]
which gives the following equation for $t_i$:
\[
    t_i = \frac{\phi S}{B} + w.
\]
This results in the following profits of the provider
\[
    \begin{split}
        \varPi_p^* &= \sum_{i=1}^m M q^* t_i - m M q^* w - \lambda_p (n - 1)\\
        &= \frac{m \phi M q^*}{B} \Bigl(A \delta - \lambda k_m - \frac{A
        b}{\max(n^\rho, k_m)^\beta} + \lambda \min(n^\rho, k_m)\Bigr) -
        \lambda_p (n - 1).
    \end{split}
\]

Note that the non-negativity of profits $\varPi_p^* \ge 0$ implies the
feasibility condition (since $\lambda_p (n - 1) \ge 0$), making the feasibility
condition redundant.

\subsection{Proof of Theorem \ref{thm:invest}}

Maximization of the provider's profits is equivalent to
\[
    \max_{n \ge 1} f(n) \defeq \lambda \min(n^\rho, k_m) - \frac{A
    b}{\max(n^\rho, k_m)^\beta} - \frac{\lambda_p B n}{m \phi M q^*}.
\]

First, we show that this function is differentiable. Notice that
\[
    \Bigl(-\frac{A b}{n^{\rho \beta}}\Bigr)' = \frac{\rho \beta A b}{n^{\rho
    \beta + 1}},\
    (\lambda n^\rho)' = \rho \lambda n^{\rho - 1}.
\]
At the point $n^\rho = k_m$, we get
\[
    \frac{\rho \beta A b}{n^{\rho \beta + 1}} = \frac{\rho \beta A
    b}{k_m^{\beta + 1}} k_m n^{-1} = \rho \lambda k_m n^{-1} = \rho \lambda
    n^{\rho - 1}.
\]
Thus, the left and right derivatives agree at $n^\rho = k_m$, and the function
is differentiable at this point. For other points, the function is
differentiable as a power function. Moreover,
\[
    f' = \rho \min\Bigl(\frac{\beta A b}{n^{\rho \beta + 1}},
    \lambda n^{\rho - 1}\Bigr) - \frac{\lambda_p B}{m \phi M q^*},
\]
and this derivative is decreasing in $n$.

Now, the necessary conditions for an extremum give
\[
    \begin{cases}
        0 = \rho \min\Bigl(\frac{\beta A b}{n^{\rho \beta + 1}},
        \lambda n^{\rho - 1}\Bigr) - \frac{\lambda_p B}{m \phi M q^*} +
        \nu,\\
        0 = \nu (n - 1),\\
        \nu \ge 0, n \ge 1.
    \end{cases}
\]
If $n = 1$, the following should hold:
\[
    \min(\beta A b, \lambda) \le \frac{\lambda_p B}{\rho m \phi M q^*}.
\]
Otherwise, $n > 1$ and we have
\[
    \min\Bigl(\frac{\beta A b}{n^{\rho \beta + 1}}, \lambda n^{\rho - 1}\Bigr)
    = \frac{\lambda_p B}{\rho m \phi M q^*}.
\]
By solving this equation, we get
\[
    n^* =
    \begin{cases}
        n_s, & n_s^\rho \le k_m\\
        n_b, & n_b^\rho \ge k_m
    \end{cases}
    = \min(n_s, n_b).
\]
where $n_s \defeq \bigl(\frac{\rho \phi m \lambda M q^*}{\lambda_p B}\bigr)^{1
/ (1 - \rho)}$, $n_b \defeq \bigl(\frac{\rho \phi m \beta A M q^* b}{\lambda_p
B}\bigr)^{1 / (1 + \rho \beta)}$.

\paragraph{Monotonicity properties}
Now we need to check the properties of the values considered. Most properties
of $n_s$ are obvious, except for its dependence on $m$ and $\gamma$ under
token-based fees. First, we show that $n_s$ increases in $m$ under token-based
fees. To show this, we note
\[
    n_s^{1 - \rho} \propto \frac{m (2 + \gamma (m - 1))}{2 + \gamma (m - 2)} =
    m + \frac{\gamma}{\frac{2 - 2 \gamma}{m} + \gamma}.
\]
Here, clearly both summands are increasing in $m$. Now, we show that $n_s$
decreases in $\gamma$ under token-based fees. We have
\[
    n_s^{1 - \rho} \propto \frac{2 M q^*}{A} = \frac{(2 - \gamma) (2 + \gamma (m
    - 1))}{2 + \gamma (m - 2)} = 2 - \frac{\gamma^2 (m - 1)}{2 + \gamma (m -
    2)}.
\]
We get
\[
    \pdv{(2 M q^* / A)}{\gamma} = \frac{- \gamma (m - 1) (4 + \gamma (m -
    2))}{(2 + \gamma (m - 2))^2} < 0.
\]

Most properties of $n_b$ are also obvious, except for the dependence on $m$ and
$\gamma$. Under subscription-based fees, we get
\[
    n_b^{1 + \rho \beta} \propto \frac{m (2 + \gamma (m - 2))}{(2 + \gamma (m -
    1))^2}.
\]
The derivative of this expression is
\[
    \frac{2 - 2 \gamma + 2 \gamma m}{(2 + \gamma (m - 1))^2} - \frac{2 m (2 +
    \gamma (m - 2)) \gamma}{(2 + \gamma (m - 1))^3} = \frac{2 (2 + \gamma (m -
    3) + \gamma^2)}{(2 + \gamma (m - 1))^3} \ge \frac{2 (1 - \gamma + \gamma
    m)}{(2 + \gamma (m - 1))^3}  > 0,
\]
proving the monotonicity. Under token-based fees, we get
\[
    n_b^{1 + \rho \beta} \propto \frac{m}{(2 + \gamma (m - 1))} =
    \frac{1}{\frac{2 - \gamma}{m} + \gamma},
\]
which is monotonically increasing in $m$. For the dependence on $\gamma$, we
note that it is mediated by $A$ under subscription-based fees and calculate the
following derivative:
\[
    - \pdv{A}{\gamma} = \frac{4 (m - 1) (2 + \gamma (m - 5) - \gamma^2 (m - 2))
    M q^*}{(2 - \gamma)^2 (2 + \gamma (m - 1))^2}.
\]
The sign of the right-hand expression is equal to the sign of $2 + \gamma (m -
5) - \gamma^2 (m - 2)$. This expression is at most a second-degree polynomial,
and it is positive at $0$ and negative at $1$, giving the desired result. Under
token-based fees, we get
\[
    n_b^{1 + \rho \beta} \propto M q^* \propto \frac{1}{2 + \gamma (m - 1)},
\]
which is decreasing in $\gamma$.

Finally, for $k_m$, most of the properties are again obvious, except the
dependence on $m$ and $\gamma$. The dependence on $\gamma$ is the same as for
$n_b$ under subscription-based fees. For $m$, we have
\[
    \frac{1}{k_m^{1 + \beta}} \propto \frac{(2 + \gamma (m - 1))^2}{2 + \gamma
    (m - 2)} = 2 + \gamma m + \frac{\gamma^2}{2 + \gamma (m - 2)}.
\]
The derivative of this expression is
\[
    \gamma - \frac{\gamma^3}{(2 + \gamma (m - 2))^2} > 0,
\]
proving the necessary result.

\subsection{Proof of Proposition \ref{prop:priv-opt}}

The proof of this proposition generally mimics the proof of
\cref{thm:priv-invest}, except for the cost sensitivity of the new objective.

\paragraph{Industry profits}
The sum of all profits results in the following expression
\[
    \begin{aligned}
        \sum_{i=1}^m & \varPi_i + \varPi_p\\
        = & \sum_i \text{Sales}_i - \text{ProductionCosts}_i -
        \text{SafetyRelatedCosts}_i - \text{FirmSafetyInvestments}_i\\
        & - \text{InferenceCosts} - \text{ProviderSafetyInvestments}\\
        \approx & \sum_i \Bigl(M (q^*)^2 - A^\text{ind} (c_i - a) - \lambda
        k_i\Bigr) - \lambda_p (n - 1),
    \end{aligned}
\]
where $A^\text{ind} \defeq \frac{2 M q^*}{2 + \gamma (m - 1)}$. Here, we used
the fact that \text{InferenceCosts} are equivalent to the per-token fee $t_i =
w$ and approximation (\ref{eq:profit-approx}).

\paragraph{Market welfare}
The consumer utility follows
\[
    u^* = \sum_{i=1}^m q_i^* - \frac{\sum_i (q_i^*)^2 + \gamma \sum_{i \neq j}
    q_i^* q_j^*}{2} - \sum_{i=1}^m p_i^* q_i^* + G.
\]

So, in the equilibrium, the total market welfare follows
\[
    W^m = M u^* + \varPi^\text{ind} = M \Bigl(\sum_{i=1}^m (1 - c_i - w) q_i^* -
    \frac{\sum_i (q_i^*)^2 + \gamma \sum_{i \neq j} q_i^* q_j^*}{2} + G\Bigr) -
    \sum_i \lambda k_i - \lambda_p (n - 1).
\]
So, the first-order approximation gives
\[
    \begin{aligned}
        W^m & \approx const + \sum_i M ((a - c_i) q^* + (1 - a - w) (q_i^* -
        q^*) - (1 + \gamma (m - 1)) q^* (q_i^* - q^*))\\
        & \quad - \sum_i \lambda k_i - \lambda_p (n - 1)\\
        & = const + \sum_i \Bigl(-A^\text{mw} (c_i - a) - \lambda k_i\Bigr) -
        \lambda_p (n - 1),
    \end{aligned}
\]
where $A^\text{mw} \defeq \frac{(3 + \gamma (m - 1)) M q^*}{2 + \gamma (m -
1)}$.

\paragraph{Private investments}
Similarly to \cref{thm:priv-invest}, we get
\[
    k^x_i = \max(0, k^x_m - n^\rho).
\]

\subsection{Proof of Proposition \ref{prop:prov-opt}}

After substituting the response $k_i(n)$, the optimal investment problem for
the provider reduces to
\[
    \max_{n} m \Bigl(\lambda \min(n^\rho, k_0) - \frac{A^x b}{\max(n^\rho,
    k_0)^\beta}\Bigr) - \lambda_p n.
\]
This equation gives the same problem as \cref{eq:provider} for $\phi \mapsto B
/ M q^*$ and $A \mapsto A^x$. Then, the proof repeats the proof of Proposition
\ref{prop:downstream} (see below). Crucially, the condition $k_0 \ge k^x_m$
ensures that the subgradient of the objective above is non-increasing in $n$.
This property reduces the number of cases arising from the objective's
non-differentiability.

\subsection{Proof of Proposition \ref{prop:downstream}}

The case $k_0 \le 1$ introduces no new constraint since $n \ge 1$.

If $k_0 \le k_m$, downstream firms' responses do not change because their objective is the same. Thus, the provider's objective and response also do not change.

Now, we consider the case $k_0 > \max(k_m, 1)$.

\paragraph{Downstream firms}
Similarly to \cref{thm:priv-invest}, we have $k = \max(0, k_0 - n^\rho)$, if
\[
    \frac{\beta A b}{\max(k_0^{\beta + 1}, n^{\rho (\beta + 1)})} - \lambda \le
    0.
\]
However, since $k_0 > k_m$, we get the following condition on the derivative of
the downstream objective:
\[
    \frac{\beta A b}{k_0^{\beta + 1}} - \lambda < 0,
\]
which implies the condition above and gives $k^* = \max(0, k_0 - n^\rho)$.

\paragraph{Provider's objective}
The solution for $k^*$ results in the following provider's objective
\[
    \max_{n \ge 1} f(n) \defeq \lambda \min(n^\rho, k_0) - \frac{A
    b}{\max(n^\rho, k_0)^\beta} - \frac{\lambda_p B n}{m \phi M q^*}.
\]
Although this function is not differentiable, its subgradient decreases in $n$ when $k_0 > k_m$. Thus, the first-order conditions would give the same answer as in \cref{thm:invest}.

\paragraph{First case, $n \le k_0^{1 / \rho}$}
We get the following Lagrange function:
\[
    L(n) = \lambda n^\rho - \frac{A b}{k_0^\beta} - \frac{\lambda_p B n}{m \phi
    M q^*} + \nu (n - 1) + \alpha (k_0^{1 / \rho} - n).
\]
The necessary conditions for an extremum give
\[
    \begin{cases}
        0 = \rho \lambda n^{\rho - 1} - \frac{\lambda_p B}{m \phi M q^*} +
        \nu - \alpha,\\
        0 = \nu (n - 1),\\
        0 = \alpha (k_0^{1 / \rho} - n),\\
        \nu \ge 0, \alpha \ge 0, n \ge 1, n \le k_0^{1 / \rho}.
    \end{cases}
\]
Hence, $n = 1$ if
\[
    0 \ge \rho \lambda - \frac{\lambda_p B}{m \phi M q^*}.
\]
$n = k_0^{1 / \rho}$ if
\[
    0 \le \rho \lambda k_0^{(\rho - 1) / \rho} - \frac{\lambda_p B}{m \phi M
    q^*}.
\]
Otherwise,
\[
    0 = \rho \lambda n^{\rho - 1} - \frac{\lambda_p B}{m \phi M q^*} \implies n
    = n_s.
\]

\paragraph{Second case, $n \ge k_0^{1 / \rho}$}
Similarly to the first case, $n = \max(1, k_0^{1 / \rho})$ if
\[
    0 \ge \frac{\rho \beta A b}{\max(k_0^{(\rho \beta + 1) / \rho}, 1)} -
    \frac{\lambda_p B}{m \phi M q^*}.
\]
Otherwise,
\[
    0 = \frac{\rho \beta A b}{n^{\rho \beta + 1}} - \frac{\lambda_p B}{m \phi M
    q^*} \implies n = n_b.
\]

\paragraph{Global solution}
Now we need to connect these cases. For the case $n \le k_0^{1 / \rho}$, we
have 3 possible values of $n$: $1$, $k_0^{1 / \rho}$, and $n_s$. For the case
$n \ge k_0^{1 / \rho}$, we have have 2 possible values: $k_0^{1 / \rho}$ and
$n_b$.

If $n = n_b$ in the second case, we should have $n = k_0^{1 / \rho}$ in the
first case, which implies that the optimal value of $n$ is $n_b$ (since
$\forall n \in [1, k_0^{1 / \rho}] \ f(n) \le f(k_0^{1 / \rho})$, but $\forall
n \ge k_0^{1 / \rho} \ f(n) \le f(n_b)$, which implies $\forall n \ge 1 \ f(n)
\le f(n_b)$). To establish this property, notice that $n = n_b$ in the second
case implies
\[
    0 < \frac{\rho \beta A b}{k_0^{(\rho \beta + 1) / \rho}} - \frac{\lambda_p
    B}{m \phi M q^*} \le \rho \lambda k_0^{(\rho - 1) / \rho} - \frac{\lambda_p
    B}{m \phi M q^*},
\]
giving $n = k_0^{1 / \rho}$ in the first case. Note that the last inequality
holds since $k_m \le k_0$.

If $n < k_0^{1 / \rho}$ in the first case, we should have $n = k_0^{1 / \rho}$
in the second case. To establish this, notice that $n < k_0^{1 / \rho}$ in the
first case implies
\[
    0 > \rho \lambda k_0^{(\rho - 1) / \rho} - \frac{\lambda_p B}{m \phi M
    q^*} \ge \frac{\rho \beta A b}{k_0^{(\rho \beta + 1) / \rho}} -
    \frac{\lambda_p B}{m \phi M q^*},
\]
giving $n = k_0^{1 / \rho}$ in the second case.

Finally, if $n = k_0^{1 / \rho}$ in both cases, we get
\[
    \rho \lambda k_0^{(\rho - 1) / \rho} \ge \frac{\lambda_p B}{m \phi M q^*}
    \ge \frac{\rho \beta A b}{k_0^{(\rho \beta + 1) / \rho}}.
\]

So, we get
\[
    n^* =
    \begin{cases}
        1, & \frac{\lambda_p B}{m \phi M q^*} \ge \rho \lambda,\\
        n_s, & \rho \lambda \ge \frac{\lambda_p B}{m \phi M q^*} \ge \rho
        \lambda k_0^{(\rho - 1) / \rho},\\
        k_0^{1 / \rho}, & \rho \lambda k_0^{(\rho - 1) / \rho} \ge
        \frac{\lambda_p B}{m \phi M q^*} \ge \frac{\rho \beta A b}{k_0^{(\rho
        \beta + 1) / \rho}},\\
        n_b, & \frac{\rho \beta A b}{k_0^{(\rho \beta + 1) / \rho}} \ge
        \frac{\lambda_p B}{m \phi M q^*},
    \end{cases}
    =
    \max(1, n_e),
\]
where
\[
    n_e =
    \begin{cases}
        n_s, & n_s \le k_0^{1 / \rho},\\
        k_0^{1 / \rho}, & n_s \ge k_0^{1 / \rho} \ge n_b,\\
        n_b, & k_0^{1 / \rho} \le n_b.
    \end{cases}
\]

\paragraph{Comparison with $n^{*, \text{old}}$}
At the same time, we have
\[
    n^{*, \text{old}} = \max(1, n_e^\text{old}), \text{ where } n_e^\text{old}
    =
    \begin{cases}
        n_s, & n_s \le k_m^{1 / \rho},\\
        n_b, & k_m^{1 / \rho} \le n_b.
    \end{cases}
\]

To show that $n^* \ge n^{*, \text{old}}$, we need to show that $n_s \ge n_b$ if
$k_m^{1 / \rho} \le n_s \le k_0^{1/\rho}$ and that $k_0^{1 / \rho} \ge n_b$ if
$k_m^{1 / \rho} \le n_b \le k_0^{1/\rho}$. The second inequality is obvious.
The first inequality follows from the following:
\[
    n_b = (n_s^{1 - \rho} k_m^{1 + \beta})^{1 / (1 + \rho \beta)} = n_s^{(1 -
    \rho) / (1 + \rho \beta)} (k_m^{1 / \rho})^{(\rho + \rho \beta) / (1 + \rho
    \beta)}
\]
Thus, $n_s \ge n_b \iff n_s \ge k_m^{1 / \rho}$, which holds in our case.

\subsection{Proof of Proposition \ref{prop:upstream}}

The regulation would not change downstream firms' action space. Thus, their
responses would not change.

The provider's final objective would not change. Hence, the derivative of the
objective will still decrease in $n$. Hence, similarly to the proof of
\cref{thm:invest}, we get that the provider will pick $n = n_0$ if
\[
    \min\biggl(\frac{\beta A b}{n_0^{\rho \beta + 1}}, \lambda n_0^{\rho -
    1}\biggr) \le \frac{\lambda_p B}{\rho m \phi M q^*}.
\]
Otherwise, the provider will pick $n_s$ or $n_b$ similarly to
\cref{thm:invest}.

\section{Alternative Provider Structures}
\label{sec:many-providers}

This section discusses how to extend our analysis to a market with several
foundation model providers, a market where another provider could enter, or a
market where firms can use open-weight models.

\paragraph{Many providers}

We suggest the following approach to extend to a market with several model
providers.

First, we add a stage that happens before everything else in which model
providers match with downstream firms. Conceptually, in this stage, model
providers pitch their model to downstream firms and make preliminary agreements
with them. A simple extension of this kind is a random assignment (e.g., the
sales department of each provider randomly passes or fails during the
negotiations with each downstream firm); more sophisticated versions might
assume some model of hidden characteristics of firms and model providers.

Second, we replace the Nash-in-Nash solution concept with the Nash-in-Nash with
Threat of Replacement solution concept \citep{h19e}. This new solution concept
changes firms' outside options. Previously, the outside option was to stop
using AI completely. Now, the firm has two outside options: it can either stop
using AI or start using a model of a different provider. The fees to the
alternative provider should be set so the alternative provider is no worse off
when it starts serving the new firm; i.e., the firm's fee under the alternative
deal should equal its ``inference costs''.

Third, the extension should also add a factor that makes switching between
providers difficult. Without such friction, providers will compete on price
until each provider receives zero surplus from serving clients and exits the
market. In other words, without such friction, the situation reduces to the
previous case with one model provider. Such switching frictions can arise from
different factors \citep{t88t}. In the first stage, firms might make
model-specific infrastructure investments that make switching costly.
Alternatively, each model provider's inference capacity may be bounded and,
hence, inference costs for a potential additional client may skyrocket, locking
them into their original model provider.

Importantly, under this extension, only each firm's outside option changes.
Hence, using the previous notation, only the parameter $\delta$ changes. Since
$\delta$ does not affect the equilibrium and optimal investment levels,
\emph{the equilibrium and optimal investment levels remain the same in the new
model}; only the fees to the provider decrease. This property suggests that the
current analysis should not qualitatively depend on the number of upstream
model providers.

\paragraph{Threat of entry}

A similar effect arises from the threat of entry. Assume that a potential
alternative provider could enter the market. The entrant's offer will again
constitute a better outside option for the downstream firms, effectively
lowering $\delta$.

Specifically, a potential competing provider will enter the market and offer
its services if the firm offers a fee that covers its inference and
data-collection costs. Formally, we assume that the entrant announces that it
will collect $n_e$ points if it enters the market. Then, we get the following
surplus for each firm (i.e., the left-hand part of \cref{eq:approx}) that
corresponds to the new outside option:
\[
    \min_{k_e} \frac{A b}{(k_e + n_e^\rho)^\beta} + \lambda k_e + \lambda_p
    (n_e - 1) - \frac{A b}{(k_i + n^\rho)^\beta} - \lambda k_i - B (t_i - w),
\]
where $k_e$ is the potential number of points a firm collects under the
alternative deal.

In this case, only a positive externality from the provider's investments
allows for non-zero provider profits. To see this, consider $n_e = n^*$; then,
for subscription-based fees, we get the following provider's profits:
\[
    \varPi_p^* = \frac{m \phi M q^*}{B} \lambda_p (n^* - 1) - \lambda_p (n^* -
    1).
\]
These profits are positive only if $m \phi M q^* / B > 1$, i.e., when $m$ is
big enough.

\paragraph{Open-weight models}
The presence of open-weight models will also only change firms' outside options
and hence will not affect investment levels. Formally, if we assume that the
open-weight model behaves as the provider's model, we get the following surplus
for each firm:
\[
    \min_{k_e} \frac{A b}{(k_e + 1)^\beta} + \lambda k_e - \frac{A b}{(k_i +
    n^\rho)^\beta} - \lambda k_i - B (t_i - w),
\]
where $k_e$ is the potential number of points collected for the open-weight
model. As in \cref{thm:priv-invest}, the cumulative equilibrium investment
level in the open-weight case is $k^*_e + 1 = \max(k_m, 1)$. This level is
generally lower than before because the provider's investments provide no
positive externality.

\section{Alternative Liability and Welfare Structures}

This section discusses how to extend our analysis to cases in which the provider is liable for safety-related accidents and how to incorporate consumer harm from such accidents into market welfare.

\subsection{Upstream Liability}
\label{sec:upstream-liability}

This section discusses how our analysis changes if we assume the upstream
provider is liable for safety-related accidents. For simplicity, we only
consider the case of token-based fees.

Structurally, we want to introduce a safety-related cost term related to
\cref{eq:provider-costs}. Using the same linearity assumption as for the term
$\text{SafetyRelatedCosts}_i$, we introduce safety-related costs of the
provider as
\begin{equation}
    \text{SafetyRelatedCosts}_p = \sum_{i = 1}^m M c_{p, i} q_i.
\end{equation}
Here, $M c_{p, i} q_i$ represents the safety-related costs arising from
accidents at firm $i$; $c_{p, i}$ is the corresponding marginal cost; and $M
q_i$ is the equilibrium quantity produced by firm $i$. To model marginal costs,
we use the same form as in \cref{eq:costs}, resulting in
\begin{equation}
    c_{p, i} = a_p + \frac{b_p}{(k_i + n^\rho)^\beta},
\end{equation}
where $a_p$ and $b_p$ are two new constants that describe the scaling of $c_{p,
i}$.

Now, \cref{eq:bargain} in the case of token-based fees require $t_i \in [w +
c_{p, i}, a_0 - c_i]$. To study this regime, we reformulate the
absence-of-disruption assumption as
\begin{equation}
    (2 - \gamma) (1 - a - a_p - w) \gg (2 + \gamma (m - 2)) (b + b_p +
    \delta'), \text{ where } \delta' \defeq a_0 - a - a_p - w.
\end{equation}
Consequently, we approximate the costs with a Taylor expansion around $\vec{c}
= (a + a_p + w, \dots, a + a_p + w)$ rather than $(a + w, \dots, a + w)$.

Additionally, we merge stages 2 and 3 of our game, resulting in Nash-in-Nash
bargaining over $t_i$ and $k_i$. We make this modification because the provider
now wants to ensure that a downstream company invests some minimal amount in
safety.

In this setting, the first-order approximation of a firm's profits does not
change significantly, but it uses $q'^* \defeq \frac{1 - a - a_p - w}{2 +
\gamma (m - 1)}$ instead of $q^*$. Additionally, we note that the provider
surplus becomes
\[
    \text{AccessFee}_i - M q_i w - M q_i c_{p, i} = M q_i (t_i - w - c_{p, i})
    \approx M q'^* (t_i - w - c_{p, i}).
\]
(Note that $\abs{t_i - w - c_{p, i}} \le \delta' + b + b_p$, which is small.)
Hence,
\[
    \varPi^*_p - \varPi^*_{p, -i} \approx M q'^* (t_i - w - c_{p, i}).
\]
And we will have the following bargaining problem:
\[
    \argmax_{k_i, t_i} \Bigl(A' \delta' - \frac{A' (b + b_p)}{(k_i +
    n^\rho)^\beta} - \lambda k_i - A' (t_i - w - c_{p, i})\Bigr)^{1 - \phi}
    (t_i - w - c_{p, i})^\phi,
\]
where $A' \defeq \frac{2 (2 + \gamma (m - 2)) M q'^*}{(2 - \gamma) (2 + \gamma
(m - 1))}$. By shifting $t_i \mapsto t_i + c_{p, i}$, we would get the same
problem as before, but with new variables $b \mapsto b + b_p, \delta \mapsto
\delta', q^* \mapsto q'^*, A \mapsto A'$.

\paragraph{Comparison with the baseline}
It is interesting to compare the baseline case to the case where one introduces
liability to the model provider. In all formulas, introducing upstream
liabilities will be reflected through the product $A b$. Previously, we had $A
b \propto (1 - a - w) b$; now, we have $A' (b + b_p) \propto (1 - a - a_p - w)
(b + b_p)$. Here we can see two effects. Directly, additional liabilities
increase safety investments through $b + b_p$. Indirectly, the additional
liabilities reduce safety investments through $(1 - a - a_p - w)$. The second
effect occurs because additional liabilities reduce the overall profitability
of all participants and therefore limit their ability to make safety
investments. Which effect dominates depends on the nature of safety-related
costs. If costs are well-optimizable (i.e., $a_p$ is small), the first effect
dominates; otherwise, the second effect dominates.

At the same time, if we assume that the regulator simply splits the liabilities
instead of introducing new ones, i.e., $a = a' + a'_p$ and $b = b' + b'_p$.
Then the equilibrium levels of investment and profits for all participants will
not change; only the fees to the model provider will increase. This happens
because the model provider might easily pass its liability to the downstream
firms through higher access fees.

\subsection{Additional Welfare Analysis}
\label{sec:welfare}

This section discusses how to incorporate consumer harm from such accidents
into market welfare.

Similar to safety-related costs, we assume that the direct harm to consumers is
proportional to production scale
\begin{equation}
    CH = \sum_{i = 1}^m M q_i c_{c, i},
\end{equation}
where $CH$ is the direct consumer harm in monetary terms (to avoid double
counting, this term only includes the harm not covered by fines paid by firms),
$M q_i$ is the production scale, and $c_{c, i}$ is the marginal increase in
harm with the production scale. Using the same logic as for \cref{eq:costs}, we
assume that
\begin{equation}
    c_{c, i} = a_c + \frac{b_c}{(k_i + n^\rho)^\beta}.
\end{equation}

Assuming $b_c / (k_i + n^\rho)^\beta$ is small, the first-order approximation
of consumer harm gives
\[
    \begin{aligned}
        CH \approx& m M q^* a_c + \sum_i M q^* (c_{c, i} - a_c) + M (q_i - q^*)
        a_c\\
        =& m M q^* a_c + \sum_i M q^* (c_{c, i} - a_c) - \frac{M a_c (c_i -
        a)}{2 + \gamma (m - 1)}\\
        =& m M q^* a_c + \sum_i \frac{M q^* b_c}{b} (c_i - a) - \frac{M q^*
        a_c}{1 - a - w} (c_i - a)
    \end{aligned}
\]

Now, we define the (complete) welfare function as
\begin{equation}
    W \defeq W^m - CH \approx const - \sum_i \Bigl(\Bigl(A^\text{mw} + \frac{M
    q^* b_c}{b} - \frac{M q^* a_c}{1 - a - w}\Bigr) (c_i - a) + \lambda
    k_i\Bigr) - \lambda_p (n - 1).
\end{equation}
Define
\begin{equation}
    A^\text{w} \defeq A^\text{mw} + \frac{M q^* b_c}{b} - \frac{M q^* a_c}{1 -
    a - w}.
\end{equation}
Then, we get
\[
    W = const - \sum_i \Bigl(A^\text{w} (c_i - a) + \lambda k_i\Bigr) -
    \lambda_p (n - 1).
\]
The welfare function has the same form as the market welfare function, but with
$A^\text{w}$ instead of $A^\text{mw}$. Thus, the optimal investment levels will
also have the same form as before.

\paragraph{Comparison with the baseline}
Interestingly, accounting for consumer harm does not necessarily lead to higher
optimal investments. Specifically, we have two effects. A direct effect,
expressed through the first additional term, suggests increasing safety
investments to reduce consumer harm. An indirect effect, expressed through the
second additional term, suggests reducing safety investments because this
action lowers quantities produced and hence reduces the number of
safety-related accidents. Which effect dominates depends on the nature of
safety-related costs. If costs are well-optimizable (i.e., $a_c$ is small), the
first effect dominates; otherwise, the second effect dominates.

\section{Safety-Related Costs Extensions}

This section considers different relaxations of the parametric form of
safety-related costs considered in the main text.

\subsection{Asymmetric Costs}
\label{sec:asym-cost}

This section relaxes the cost-symmetry assumption.

We consider asymmetric cost functions
\begin{equation}
    c_i = a_i + \frac{b_i}{(k_i + n^\rho)^\beta}
\end{equation}
and allow asymmetric initial costs $a_{0, i}$. In this setting, we reformulate
the absence-of-disruption assumption as
\begin{equation}
    \forall i \, (2 - \gamma) (1 - a_i - w) + \gamma \sum_j (a_j - a_i) \gg (2
    + \gamma (m - 2)) (b_i + \delta_i), \text{ where } \delta_i \defeq a_{0, i}
    - a_i - w.
\end{equation}

This assumption will again allow us to use the first-order approximation and
get the following asymmetric versions of $A$ and $q^*$:
\begin{equation}
    A_i \defeq \frac{2 (2 + \gamma (m - 2)) M q^*_i}{(2 - \gamma) (2 + \gamma
    (m - 1))}, \ q^*_i \defeq \frac{(2 - \gamma) (1 - w) - (2 + \gamma (m - 2))
    a_i + \gamma \sum_{j \neq i} a_j}{(2 - \gamma) (2 + \gamma (m - 1))}.
\end{equation}
And we get the following equilibrium private investments
\begin{equation}
    k^*_i = \max(0, k_{m, i} - n^\rho), \ k_{m, i} \defeq (\beta b_i A_i /
    \lambda)^{1 / (1 + \beta)}.
\end{equation}
These results give the following provider's profits:
\begin{equation}
    \varPi_p^* = \sum_i \frac{\phi M q^*_i}{B_i} (A_i \delta_i - \lambda k_{m,
    i} - \frac{A_i b_i}{\max(n^\rho, k_{m, i})^\beta} + \lambda \min(n^\rho,
    k_{m, i})) - \lambda_p (n - 1),
\end{equation}
where $B_i = M q^*_i$ for subscription-based fees and $B_i = A_i$ for
token-based fees.

The provider's profit function is differentiable, and one can find the
provider's investment level by using the first-order conditions. However, this
solution is hard to interpret and analyze. To get interpretable expressions,
consider two limiting cases: $(n^*)^\rho \le \min_i k_{m, i}$ and $(n^*)^\rho
\ge \max_i k_{m, i}$. These cases correspond to the cases $n^* = \max(1, n_s)$
and $n^* = \max(1, n_b)$ in the main text. In the first case, we get
\[
    n^* = \max\Bigl(1, \Bigl(\frac{\phi M \rho \lambda}{\lambda_p} \sum_i
    \frac{q^*_i}{B_i}\Bigr)^{1 / (1 - \rho)}\Bigr).
\]
This expression resembles $n_s$. However, instead of symmetric ratios $q^* /
B$, we sum their asymmetric analogs. In the second case, we get
\[
    n^* = \max\Bigl(1, \Bigl(\frac{\phi M \rho \beta}{\lambda_p} \sum_i
    \frac{b_i A_i q^*_i}{B_i}\Bigr)^{1 / (1 + \rho \beta)}\Bigr).
\]
We again see that the result resembles $n_b$.

\subsection{Different Form of Costs}
\label{sec:diff-cost}

This section relaxes the requirement that the cost follows the parametric form
(\ref{eq:costs}). Specifically, we assume that
\begin{equation}
    c_i = a + f(k_i + n^\rho),
\end{equation}
where $f$ is differentiable, monotonically decreasing, and convex. In this
setting, the absence-of-disruption assumption becomes
\begin{equation}
    (2 - \gamma) (1 - a - w) \gg (2 + \gamma (m - 2)) (f(1) + \delta).
\end{equation}
This assumption will again allow us to use the first-order approximation.

We get the following equilibrium private investments:
\begin{equation}
    k^* = \max(0, k_m - n^\rho), \text{ where } -f'(k_m) = \frac{\lambda}{A}.
\end{equation}
Since $f$ is convex, $-f'$ is monotonically decreasing; hence, the behavior of
$k_m$ is similar to its behavior in the main text. We get the following
provider's profits:
\[
    \varPi_p^* = \frac{m \phi M q^*}{B} \Bigl(A \delta - \lambda k_m - A
    f(\max(n^\rho, k_m)) + \lambda \min(n^\rho, k_m)\Bigr) - \lambda_p
    n + \lambda_p.
\]
This expression results in the following first-order condition for the
equilibrium investments:
\begin{equation}
    \min(\lambda \rho n^{\rho - 1}, -A \rho n^{\rho - 1} f'(n^\rho)) =
    \frac{\lambda_p B}{m \phi M q^*}.
\end{equation}
Thus, we get $n^* = \max(1, \min(n_s, n_b))$ as before. Moreover, we get the
same expression for $n_s$, while $n_b$ should satisfy
\begin{equation}
    -n_b^{\rho - 1} f'(n_b^\rho) = \frac{\lambda_p B}{\rho \phi m A M q^*}.
\end{equation}
Since $-n^{\rho - 1} f'(n^\rho)$ is monotonically decreasing, the behavior of
$n_b$ is similar to its behavior in the main text.

\subsection{Multi-Objective Costs}
\label{sec:many-cost}

This section discusses how to modify the framework to allow optimizing for
several properties that follow polynomial scaling laws. Assume we have $R$
properties that influence costs, and that the costs associated with these
properties grow linearly with production volume. The results of \citet{s25gen}
motivate the following structure of marginal cost
\begin{equation}
    c_i = a + \sum_{r = 1}^R \frac{b_r}{(k_{r, i} + n_r^{\rho_r})^{\beta_r}},
\end{equation}
where $r$ is the property, $b_r$, $\rho_r$, and $\beta_r$ are scaling constants
associated with this property, and $k_{r, i}$ and $n_r$ are firm $i$'s and
provider's investments in property $r$. In this case, we can assume the
following version of the absence-of-disruption assumption:
\begin{equation}
    (2 - \gamma) (1 - a - w) \gg (2 + \gamma (m - 2)) (\delta + \sum_r b_r).
\end{equation}

Since the cost structure is separable across properties and we use a
first-order approximation, maximizing profits across investments in all
properties reduces to maximizing profits for each property individually Hence,
we get the same results as before for each property, and our results do not
change qualitatively.

\section{Extensions}
\label{sec:extensions}

This appendix discusses possible extensions to our framework and theoretical
model.

\subsection{Nonlinear Inference Costs}
\label{sec:nonlinear-inference}

This section informally discusses how to incorporate convex nonlinear inference
costs into our analysis. In this section, we assume that nonlinearity is small
in some sense and clarify what it means below. For simplicity, we focus on
token-based fees.

This section again uses the first-order approximation. However, the zero-order
approximation of $q_i$ is no longer straightforward because the inference costs
are nonlinear. To find this approximation, assume that $q^*$ and $t^*$ are good
approximations for $q_i$ and $t_i$. Then, we should have the following
provider's surplus:
\[
    \begin{aligned}
        \varPi^*_p - \varPi^*_{p, -i} & \\
        = & \sum_j M q_j t_j - \sum_{j \neq i} M q_{j, -i} t_j - C(\sum_j q_j)
        + C(\sum_{j \neq i} q_j)\\
        \approx & m M q^* t^* + \sum_j (M q^* (t_j - t^*) + M (q_j - q^*)
        t^*)\\
        & - (m - 1) M q^* t^* - \sum_{j \neq i} (M q^* (t_j - t^*) + M (q_{j,
        -i} - q^*) t^*)\\
        & - C(m q^*) - C'(m q^*) \sum_j (q_j - q^*)\\
        & + C((m - 1) q^*) + C'((m - 1) q^*) \sum_{j \neq i} (q_{j, -i} - q^*)
    \end{aligned}
\]
Since the first-order expression bounds firms' surplus, the provider's surplus
should also be first-order. This logic gives
\[
    M q^* t^* - C(m q^*) + C((m - 1) q^*) = 0 \implies t^* = \frac{C(m q^*) -
    C((m - 1) q^*)}{M q^*}.
\]
Now, the expression for $q^*$ is
\[
    q^* = \frac{1 - a - t^*}{2 + \gamma (m - 1)}.
\]
Substituting $t^*$ into this expression gives an equation for $q^*$.

Now we proceed assuming that $q^*$ and $t^*$ are some fixed expressions. We get
the following provider's surplus:
\[
    \begin{aligned}
        \varPi^*_p - \varPi^*_{p, -i} \approx & M q^* (t_i - t^*) + \sum_j M
        (q_j - q^*) t^* - \sum_{j \neq i} M (q_{j, -i} - q^*) t^*\\
        & - C'(m q^*) \sum_j (q_j - q^*) + C'((m - 1) q^*) \sum_{j \neq i}
        (q_{j, -i} - q^*).
    \end{aligned}
\]
To make this expression tractable, we assume that $C'(m q^*) - M t^*$ and
$M t^* - C'((m - 1) q^*)$ are small. (Even without this assumption, it is
possible to proceed with the analysis. However, the resulting expression will
differ from the case analyzed in the main text.) These two expressions
precisely control the strength of nonlinearity of $C$ on the interval $[(m - 1)
q^*, m q^*]$. This assumption gives
\[
    \varPi^*_p - \varPi^*_{p, -i} \approx M q^* (t_i - t^*).
\]
Then, we can proceed with our analysis as before.

In summary, we make the following assumptions. First, we assume that $C$ is
convex and monotonically increasing. Then, we solve the system
\begin{equation}
    \begin{cases}
        t^* = \frac{C(m q^*) - C((m - 1) q^*)}{M q^*},\\
        q^* = \frac{1 - a - t^*}{2 + \gamma (m - 1)}.
    \end{cases}
\end{equation}
Then, we assume that the nonlinearity is small
\begin{equation}
    q^* \gg C'(m q^*) - M t^* + M t^* - C'((m - 1) q^*)  = C'(m q^*) - C'((m -
    1) q^*).
\end{equation}
Finally, we make the following absence-of-disruption assumption:
\begin{equation}
    (2 - \gamma) (1 - a - t^*) \gg (2 + \gamma (m - 2)) (\delta + b), \text{
        where } \delta \defeq a_0 - a - t^*.
\end{equation}
Under these assumptions, we get the following approximation for
\cref{eq:bargain}:
\begin{equation}
    \argmax_{t_i} \Bigl(A \delta - \frac{A b}{(k_i + n^\rho)^\beta} - \lambda
    k_i - A (t_i - t^*)\Bigr)^{1 - \phi} (t_i - t^*)^\phi,
\end{equation}
where $A$ is defined as previously, but using the new $q^*$. The analysis of
this expression mirrors the analysis presented in the main text.

\subsection{Disruption}
\label{sec:disruption}

This section informally discusses possible relaxations of the
absence-of-disruption assumption and how the analysis changes without this
assumption.

\paragraph{Bound on $b$}
To relax the bound on $b$, one can show that investments will always exceed
some minimum level $k_\text{min}$. Then one needs to upper-bound $b /
(k_\text{min})^\beta$ instead of $b$. To establish a minimum level, one can use
a ``bootstrap'' procedure. First, one considers the ``worst case of costs'',
$c^\text{eff}_i = a + w$, which incentivizes safety investments the least, and
finds the minimum level of investments in this worst case, $k_{\text{min}, 1}$.
(To show that the considered case is the worst case, one can compute the second
derivatives of $\varPi^{m, *}_i$ over $\vec{k}$.) Using this number, one now
bounds the investments from above and gets $k_{\text{max}, 1}$. The argument is
then repeated using the bound implied by $k_{\text{max}, 1}$ and so forth.
After sufficiently many iterations, this yields the desired minimum investment
level.

\paragraph{Bound on $\delta$}
An analytic expression is possible in the opposing limiting case, where
$\delta$ is large, and the company exits the market if it does not adopt the AI
technology. In this case, the outside option in the Nash bargaining problem
corresponds to zero profit. In this case, the firms' investment response
functions remain the same (since a firm's surplus still depends on $k_i$ in the
same way). However, the model provider's investment would change.

\paragraph{Allowing disruption}
Now, we discuss cases in which firms can drive competitors out of the market.
First, if it is easy to make certain firms exit by simply adopting the
technology (e.g., costs improve significantly with AI for some firms, but the
improvement is small for others, making them uncompetitive), then the analysis
mirrors our current analysis but with fewer firms. If, on the other hand, it is
possible to make some firms exit the market, but it requires investment to
exceed a threshold, then firms' incentives will exhibit a kink (a point where
the derivative jumps). Below and above the threshold, the firms' profits remain
smooth and follow our analysis. However, crossing the threshold causes some
firms to exit, creating a kink in the marginal value of investment (profits
will remain continuous). Hence, the investment levels can 1) continue to follow
\cref{thm:priv-invest}, 2) lie at the kink, or 3) fall to zero if the firm
exits. The precise structure of these kinks depends on the asymmetry
assumptions and is therefore difficult to predict.

\clearpage
\section*{NeurIPS Paper Checklist}

\begin{enumerate}

\item {\bf Claims}
    \item[] Question: Do the main claims made in the abstract and introduction accurately reflect the paper's contributions and scope?
    \item[] Answer: \answerYes{}
    \item[] Justification: The description of the framework is based on Section
        3. The description of the results reflect Sections 5 and 6.

\item {\bf Limitations}
    \item[] Question: Does the paper discuss the limitations of the work performed by the authors?
    \item[] Answer: \answerYes{}
    \item[] Justification: We explicitly discuss limitations in Section 7.

\item {\bf Theory assumptions and proofs}
    \item[] Question: For each theoretical result, does the paper provide the full set of assumptions and a complete (and correct) proof?
    \item[] Answer: \answerYes{}
    \item[] Justification: Appendix contains the proofs of all results.

    \item {\bf Experimental result reproducibility}
    \item[] Question: Does the paper fully disclose all the information needed to reproduce the main experimental results of the paper to the extent that it affects the main claims and/or conclusions of the paper (regardless of whether the code and data are provided or not)?
    \item[] Answer: \answerNA{} 
    \item[] Justification: The paper is theoretical.

\item {\bf Open access to data and code}
    \item[] Question: Does the paper provide open access to the data and code, with sufficient instructions to faithfully reproduce the main experimental results, as described in supplemental material?
    \item[] Answer: \answerNA{} 
    \item[] Justification: The paper is theoretical.

\item {\bf Experimental setting/details}
    \item[] Question: Does the paper specify all the training and test details (e.g., data splits, hyperparameters, how they were chosen, type of optimizer) necessary to understand the results?
    \item[] Answer: \answerNA{} 
    \item[] Justification: The paper is theoretical.

\item {\bf Experiment statistical significance}
    \item[] Question: Does the paper report error bars suitably and correctly defined or other appropriate information about the statistical significance of the experiments?
    \item[] Answer: \answerNA{} 
    \item[] Justification: The paper is theoretical.

\item {\bf Experiments compute resources}
    \item[] Question: For each experiment, does the paper provide sufficient information on the computer resources (type of compute workers, memory, time of execution) needed to reproduce the experiments?
    \item[] Answer: \answerNA{} 
    \item[] Justification: The paper is theoretical.
    
\item {\bf Code of ethics}
    \item[] Question: Does the research conducted in the paper conform, in every respect, with the NeurIPS Code of Ethics \url{https://neurips.cc/public/EthicsGuidelines}?
    \item[] Answer: \answerYes{} 
    \item[] Justification: The authors have read and follow the Code of Ethics.

\item {\bf Broader impacts}
    \item[] Question: Does the paper discuss both potential positive societal impacts and negative societal impacts of the work performed?
    \item[] Answer: \answerNA{} 
    \item[] Justification: We do not foresee any direct negative impact of our
        work. On the other hand, the positive impact of our work is hard to
        evaluate.
    
\item {\bf Safeguards}
    \item[] Question: Does the paper describe safeguards that have been put in place for responsible release of data or models that have a high risk for misuse (e.g., pre-trained language models, image generators, or scraped datasets)?
    \item[] Answer: \answerNA{} 
    \item[] Justification: The paper is theoretical.

\item {\bf Licenses for existing assets}
    \item[] Question: Are the creators or original owners of assets (e.g., code, data, models), used in the paper, properly credited and are the license and terms of use explicitly mentioned and properly respected?
    \item[] Answer: \answerNA{} 
    \item[] Justification: The paper is theoretical.

\item {\bf New assets}
    \item[] Question: Are new assets introduced in the paper well documented and is the documentation provided alongside the assets?
    \item[] Answer: \answerNA{} 
    \item[] Justification: The paper is theoretical.

\item {\bf Crowdsourcing and research with human subjects}
    \item[] Question: For crowdsourcing experiments and research with human subjects, does the paper include the full text of instructions given to participants and screenshots, if applicable, as well as details about compensation (if any)? 
    \item[] Answer: \answerNA{} 
    \item[] Justification: The paper is theoretical.

\item {\bf Institutional review board (IRB) approvals or equivalent for research with human subjects}
    \item[] Question: Does the paper describe potential risks incurred by study participants, whether such risks were disclosed to the subjects, and whether Institutional Review Board (IRB) approvals (or an equivalent approval/review based on the requirements of your country or institution) were obtained?
    \item[] Answer: \answerNA{} 
    \item[] Justification: The paper is theoretical.

\item {\bf Declaration of LLM usage}
    \item[] Question: Does the paper describe the usage of LLMs if it is an important, original, or non-standard component of the core methods in this research? Note that if the LLM is used only for writing, editing, or formatting purposes and does \emph{not} impact the core methodology, scientific rigor, or originality of the research, declaration is not required.
    \item[] Answer: \answerNA{} 
    \item[] Justification: The authors used LLMs only for text editing.

\end{enumerate}

\end{document}